\documentclass{article} 
\usepackage{iclr2026_conference,times}
\usepackage{hyperref}
\usepackage{url}
\usepackage{booktabs}
\usepackage{multirow}
\usepackage{siunitx}  
\usepackage{graphicx}
\usepackage{xcolor}
\usepackage{pgfplots}
\pgfplotsset{compat=1.18}

\usepackage{amsmath,amsfonts,bm}

\def\eqref#1{equation~\ref{#1}}

\def\1{\bm{1}}

\DeclareMathAlphabet{\mathsfit}{\encodingdefault}{\sfdefault}{m}{sl}
\SetMathAlphabet{\mathsfit}{bold}{\encodingdefault}{\sfdefault}{bx}{n}

\usepackage{hyperref}
\usepackage{url}

\title{Is continuous CoT better suited for multi-lingual reasoning?}

\author{Ali Hamza Bashir\textsuperscript{1, 2}, 
Behzad Shomali\textsuperscript{1, 3}, 
Markus Frey\textsuperscript{1, 2, 3}, 
Mehdi Ali\textsuperscript{1, 2},
Rafet Sifa\textsuperscript{1, 2, 3}\\
\& \textbf{David Berghaus\textsuperscript{1, 2}}\\
Lamarr Institute\textsuperscript{1}, Fraunhofer IAIS\textsuperscript{2}, University of Bonn\textsuperscript{3}
\\
\small \texttt{david.berghaus@iais.fraunhofer.de} 
}

\iclrfinalcopy 
\begin{document}

\maketitle

\begin{abstract}
We investigate whether performing reasoning in a continuous latent space leads to more robust multilingual capabilities. We compare Continuous Chain-of-Thought (using the CODI framework) against standard supervised fine-tuning across five typologically diverse languages: English, Chinese, German, French, and Urdu. Our experiments on GSM8k and CommonsenseQA demonstrate that continuous reasoning significantly outperforms explicit reasoning on low-resource languages, particularly in zero-shot settings where the target language was not seen during training. Additionally, this approach achieves extreme efficiency, compressing reasoning traces by approximately $29\times$ to $50\times$. These findings indicate that continuous latent representations naturally exhibit greater language invariance, offering a scalable solution for cross-lingual reasoning.
\end{abstract}

\section{Introduction \& Related Work}
While LLMs have demonstrated impressive reasoning capabilities, their performance varies dramatically across languages. Low-resource languages in particular suffer from substantially degraded results compared to high-resource languages like English \citep{lai2024mcot, barua2025long, liu2026large}. This disparity poses a fundamental challenge: how can we enable more equitable reasoning performance across linguistically diverse settings?

Existing approaches have explored several strategies to address this challenge. One common approach translates multilingual prompts into a high-resource pivot language like English before applying chain-of-thought (CoT) reasoning \citep{shi2022language, ahuja-etal-2023-mega}. However, this creates a bottleneck at the translation step, where subtle linguistic nuances may be lost \citep{nicholas2023lost}. More recent work has taken a different direction by directly fine-tuning LLMs on multilingual CoT data \citep{lai2024mcot, barua2025long}, showing improvements for low-resource languages. Yet this approach faces scalability issues (covering hundreds of languages during fine-tuning is impractical) and risks catastrophic forgetting as more languages are added.

We explore a different angle: what if the reasoning process itself could operate in a more language-agnostic representation space? Recent advances in continuous latent reasoning \citep{coconut,shen2025codicompressingchainofthoughtcontinuous} offer a promising direction. Rather than verbalizing each reasoning step in natural language tokens, these methods compress reasoning into continuous hidden representations. This approach is motivated by classic findings showing that word embeddings across different languages tend to occupy similar geometric spaces \citep{mikolov2013exploiting}. If reasoning happens in continuous space rather than through explicit language tokens, the representations might naturally exhibit greater language invariance.

\section{Setup}
To investigate whether continuous latent reasoning enables more language-agnostic representations, we conduct experiments across 5 typologically diverse languages: \textbf{English}, \textbf{Chinese}, \textbf{German}, \textbf{French}, and \textbf{Urdu}. This selection spans diverse branches diverse branches (Germanic, Romance, Indo-Iranian) and families (Sino-Tibetan) \citep{ethnologue} and writing systems (Latin, Chinese characters, Arabic script), providing a rigorous test of cross-lingual generalization.

We compare the base model with two fine-tuning strategies: (i) Classical CoT-SFT (see section \ref{sec:cot_sft}) (ii) Continuous CoT using the training recipe of \citep{shen2025codicompressingchainofthoughtcontinuous} (see section \ref{sec:codi}). Both strategies are used in various multilingual configurations (see sections \ref{sec:english_only}-\ref{sec:mixture_of_languages}) to investigate which strategy achieves the best performance among different languages.

We use the same hyperparameters as \citep{shen2025codicompressingchainofthoughtcontinuous} and keep them constant for all configurations to ensure comparability between both fine-tuning approaches. Complete hyperparameter details are provided in Appendix~\ref{app:hyperparameters}

\subsection{Data}
\label{sec:data}
\subsubsection{Datasets}
We use the following datasets to perform our experiments:

\paragraph{GSM8k-Aug-NL} This dataset is based on the GSM8k grade-school math reasoning benchmark \citep{cobbe2021trainingverifierssolvemath}. We use the augmented version from \citep{deng2023implicit}, which expands the original 7.5k training examples to 385k training samples and 1.3k test samples and contains CoT reasoning traces in natural language.

\paragraph{CommonsenseQA-CoT} Derived from CommonsenseQA \citep{talmor-etal-2019-commonsenseqa}, a multiple-choice question answering dataset requiring commonsense reasoning. Since the original dataset lacks CoT annotations, we’ve used the annotations provided by \cite{shen2025codicompressingchainofthoughtcontinuous}. This results in approximately 8.1k training and 1.2k validation examples.

\subsubsection{Multilingual Data Construction}
\label{sec:data_construction}

For GSM8k-Aug-NL, we create multilingual versions by translating questions into our target languages while preserving mathematical expressions and CoT structure. We employ \texttt{Llama-3.3-70B-Instruct}~\citep{touvron2023llama} for German and French, \texttt{Qwen2.5-72B-Instruct}~\citep{qwen2024tech} for Chinese, and \texttt{GPT-5-mini} for Urdu, all with low temperature (0.1) to ensure consistent translations. Translation guidelines include: (i) preserving all numerical values exactly, (ii) keeping mathematical expressions unchanged, (iii) adapting person names to culturally appropriate alternatives in target languages, and (iv) ensuring natural, native-sounding output. Samples with inadequate translation quality were discarded. For CommonsenseQA, we use \texttt{GPT-5-mini} for all language translations. All Urdu translations were verified and corrected by a native speaker. To enable rigorous cross-lingual evaluation, we ensure zero overlap between languages meaning that each unique problem appears in exactly one language, preventing any data leakage during training. Dataset statistics are provided in Appendix~\ref{app:dataset_construction}.

\subsection{Model}
\subsubsection{Base Architecture}
We build upon \texttt{LLaMA3.2-1B-Instruct} \citep{grattafiori2024llama} as our base language model. This model officially supports 8 different languages \citep{grattafiori2024llama}, but since it has been pre-trained on a large-scale web corpus containing numerous languages, it often performs reasonably well for languages that are not officially supported. This becomes apparent in table \ref{tab:appendix_base_gsm8k_aug_by_language} where the base model achieves similar performance in German \& French (which are officially supported) and Chinese (which is not officially supported).

\subsubsection{Continuous Latent Reasoning via CODI}
\label{sec:codi}
To enable reasoning in continuous latent space, we employ the CODI training framework \citep{shen2025codicompressingchainofthoughtcontinuous}. CODI trains a single shared model to jointly optimize two distinct reasoning modes: explicit token-based reasoning (teacher task) and implicit continuous reasoning (student task). This self-distillation setup allows the model to compress verbose natural language reasoning steps into compact continuous representations while preserving reasoning capability.

\paragraph{Teacher Task: Explicit CoT Reasoning}
The teacher task learns standard chain-of-thought generation by predicting a sequence of explicit reasoning tokens $c$ followed by the final answer $y$. The training objective is a standard cross-entropy loss:
\begin{equation}
    \label{eq:sft_loss}
    \mathcal{L}_{\text{teacher}} = -\frac{1}{N} \sum_{i=1}^{N} \log P(r_i \mid r_{1:i-1}, Q),
\end{equation}
where $Q$ represents the input question, $r = [c, y]$ is the concatenated sequence of CoT and answer tokens, and $N$ is the total sequence length.

\paragraph{Student Task: Continuous Thought Generation}
The student task generates reasoning in a continuous latent space by autoregressively propagating hidden states $Z = \{z_1, \ldots, z_K\}$ between learnable \texttt{<bot>} and \texttt{<eot>} tokens. An MLP projects these states between steps to distinguish latent reasoning from token embeddings. The objective is to predict the answer $y$ given these thoughts:
\begin{equation}
    \mathcal{L}_{\text{student}} = - \frac{1}{M} \sum_{i=1}^{M} \log P(y_i \mid y_{1:i-1}, Q, Z).
\end{equation}

\paragraph{Knowledge Distillation Mechanism}
To transfer reasoning capabilities and prevent the continuous representations from drifting too far from the initial language representations, CODI aligns the student's hidden activations with the teacher's at the token immediately preceding the answer (e.g., “:”). This anchors the latent reasoning to the explicit trace via an L1 loss:
\begin{equation}
    \mathcal{L}_{\text{KD}} = \frac{1}{L} \sum_{l=1}^{L} \|\text{sg}[\mathbf{h}_{\text{teacher}}^l] - \mathbf{h}_{\text{student}}^l\|_1,
\end{equation}
\noindent where $\text{sg}[\cdot]$ is a stop-gradient ensuring unidirectional knowledge transfer.

\paragraph{Training Objective}
The complete training loss combines all three components:
\begin{equation}
    \mathcal{L} = \alpha \mathcal{L}_{\text{student}} + \beta \mathcal{L}_{\text{KD}} + \gamma \mathcal{L}_{\text{teacher}},
\end{equation}

\subsubsection{CoT-SFT Baseline}
\label{sec:cot_sft}
As a baseline, we use \eqref{eq:sft_loss} to train a standard Chain-of-Thought Supervised Fine-Tuning (CoT-SFT). Note that \eqref{eq:sft_loss} excludes the prompt in the training objective, meaning that the model only receives gradients for predicting the reasoning trace.

\section{Experiments}

\subsection{Training on English only}
\label{sec:english_only}
To establish a baseline performance, we train both CODI and CoT-SFT on English data (meaning that both the question and the CoT traces are in English). Our results can be found in table \ref{tab:english_only_training}. Similar to \citep{shen2025codicompressingchainofthoughtcontinuous} we found that CODI performs worse on \textsc{GSM8k} and better on \textsc{CommonsenseQA} when evaluated in English. Moreover, in line with \citep{barua2025long,lai2024mcot} we found that the performance gets significantly worse for both models in other languages. CODI performs however better than CoT-SFT in low-resource languages for both datasets.

\begin{table}[t]
    \centering
    \small
    \sisetup{table-format=2.2}
    \caption{Accuracy (\%) on GSM8K-AUG-NL and CommonsenseQA (test splits) comparing CODI vs.\ CoT-SFT when trained on English.}
    \label{tab:english_only_training}
    \begin{tabular}{l SS SS}
        \toprule
        \multirow{2}{*}{\textbf{Test Language}} & 
        \multicolumn{2}{c}{\textbf{GSM8K-AUG-NL}} & 
        \multicolumn{2}{c}{\textbf{CommonsenseQA}} \\
        \cmidrule(lr){2-3} \cmidrule(lr){4-5}
        & {\textbf{CODI}} & {\textbf{CoT-SFT}} & {\textbf{CODI}} & {\textbf{CoT-SFT}} \\
        \midrule
        German  & 31.79 & 36.09 & 58.48 & 38.41 \\
        French  & 32.35 & 38.21 & 57.66 & 46.03 \\
        Chinese & 30.46 & 27.37 & 57.58 & 48.40 \\
        English & 46.08 & 55.04 & 71.42 & 63.23 \\
        Urdu    & 9.58  & 5.38  & 32.57 & 21.87 \\
        \bottomrule
    \end{tabular}

    \vspace{0.35em}
    \begin{flushleft}
    \footnotesize\textbf{Note:} CODI preprocessing removes the final sentence from reference CoTs. In GSM8K (English train), \textbf{28.30\%} of CoTs contain only one sentence, resulting in empty references which disadvantages CODI on that specific dataset.
    \end{flushleft}
\end{table}

\subsection{Training on a Mixture of Languages}
\label{sec:mixture_of_languages}
To improve the multilingual performance we investigate both CODI and CoT-SFT in two multilingual training configurations. In the first setup, we train on English, German, French, and Chinese, deliberately excluding Urdu to assess zero-shot generalization to a low-resource language. In the second setup, we include Urdu during training to examine performance when the low-resource language is present in the training data. 

\begin{figure}[t]
    \centering
    \includegraphics[width=1.0\linewidth]{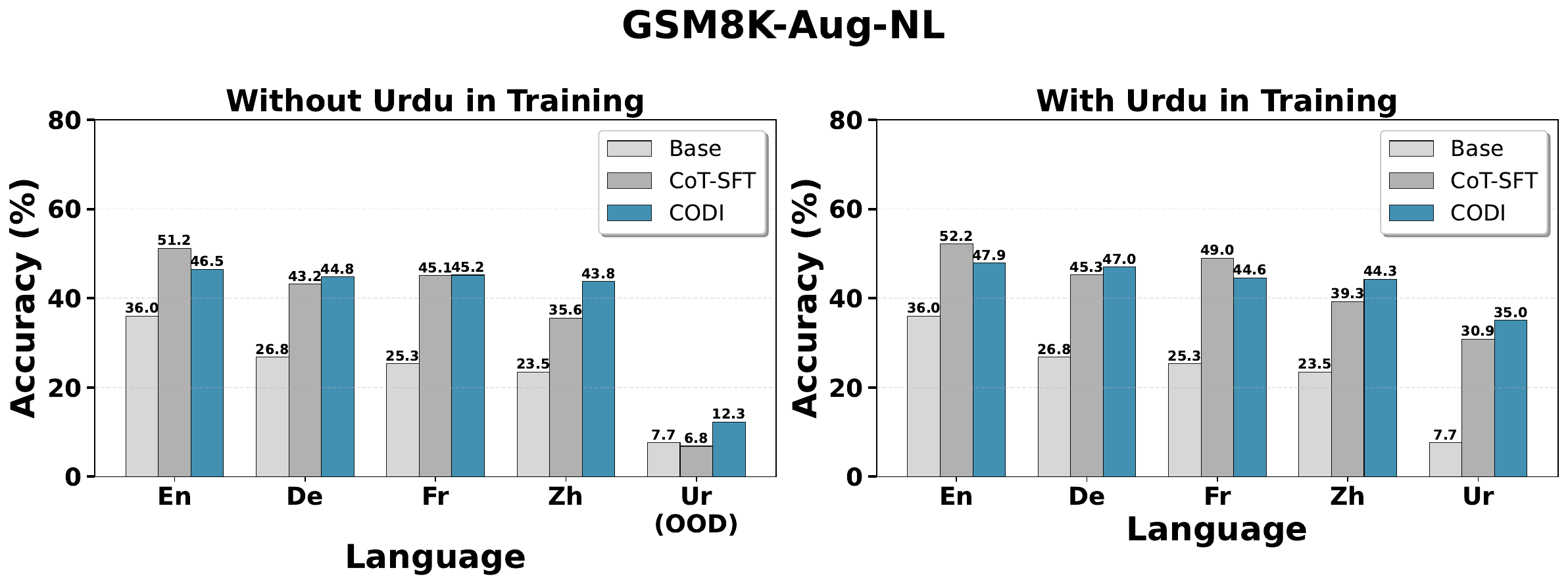}
    \vspace{-0.5em} 
    \caption{Performance of \texttt{LLaMA3.2-1B-Instruct} trained on multi-lingual GSM8k-Aug-NL data. Both models perform similarly, with CoT-SFT performing better for high-resource languages and CODI performing better for low-resource languages. Moreover, CODI performs significantly better than CoT-SFT on Urdu when it is Out-of-distribution (OOD) (i.e., when Urdu was not part of the fine-tuning data).}
    \label{fig:gsm8k_aug_nl}
    
    \vspace{1em} 
    
    \includegraphics[width=1.0\linewidth]{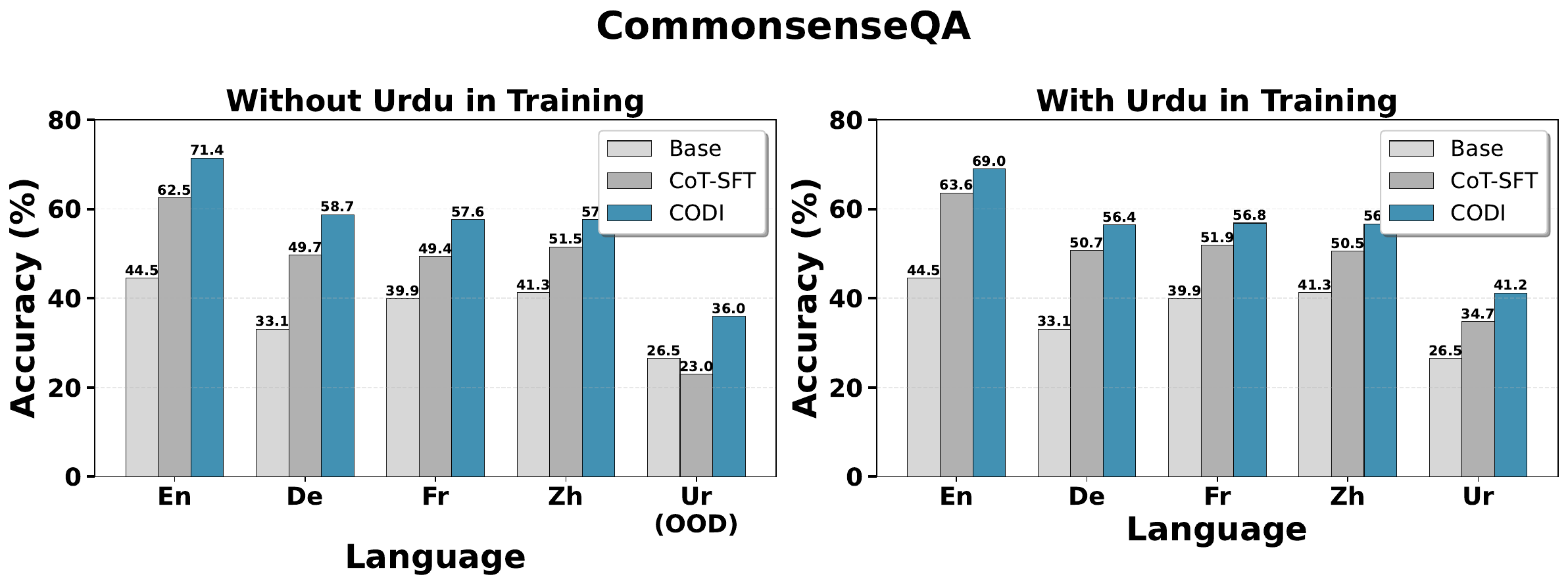}
    \vspace{-0.5em} 
    \caption{Performance of \texttt{LLaMA3.2-1B-Instruct} trained on multi-lingual CommonsenseQA data. CODI performs better across all languages.}
    \label{fig:commonsense_qa}
\end{figure}

Both approaches substantially outperform the base model \texttt{LLaMA3.2-1B-Instruct} across all languages and also improve the multilingual performance compared to english-only training, as evident in figures \ref{fig:gsm8k_aug_nl} \& \ref{fig:commonsense_qa}. Moreover, contrary to the single-language results in section \ref{sec:english_only} (and the results reported in \citep{shen2025codicompressingchainofthoughtcontinuous}) where CODI performs worse on \textsc{GSM8k}, we find that CODI matches the performance of classical CoT-SFT much better, despite using significantly less output tokens (see section \ref{sec:number_of_cot_tokens}). To be more precise, our results for \textsc{GSM8k} show that CODI performs noticeably better than COT-SFT for low-resource languages and slightly worse for high-resource languages. For \textsc{CommonsenseQA}, CODI performs better among all languages.

\begin{figure}
    \centering
    \includegraphics[width=0.75\linewidth]{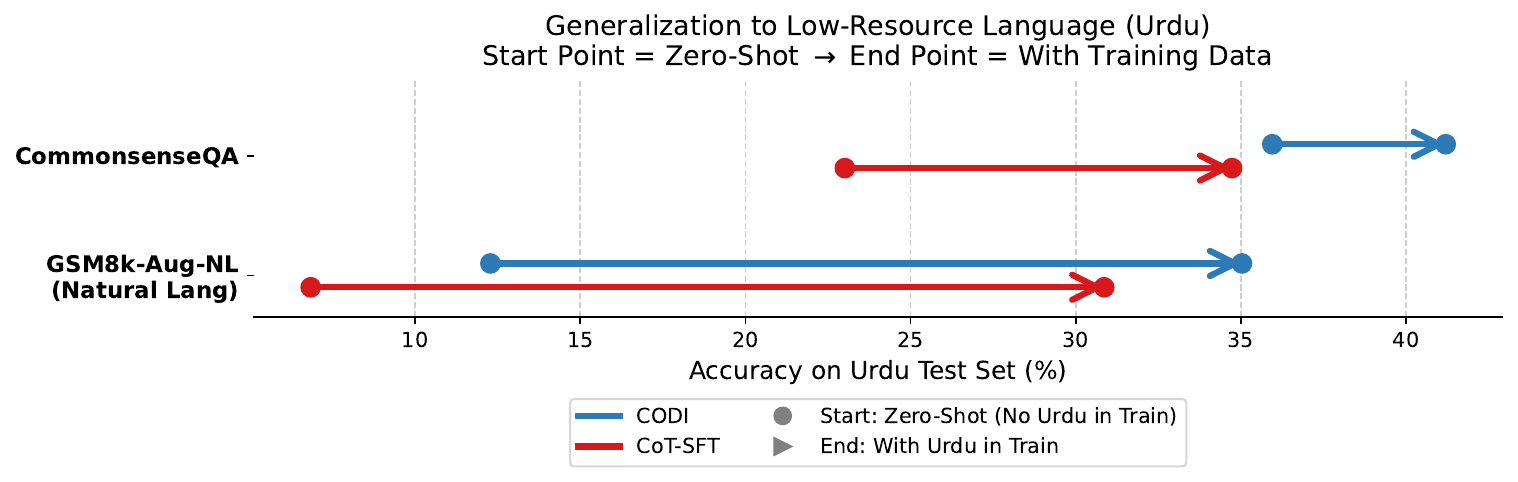}
    \caption{Performance on a low-resource language (Urdu). The latent reasoning approach works significantly better than CoT-SFT in all cases.}
    \label{fig:urdu_fig}
\end{figure}

Remarkably, CODI performs significantly better on Urdu even if it has not been trained on it during fine-tuning (see figure \ref{fig:urdu_fig}). To emphasize, even when Urdu is out-of-distribution, CODI performs better on \textsc{CommonsenseQA} (35.95\%) than CoT-SFT which had Urdu in the fine-tuning data (34.73\%).
This indicates that the latent thinking of CODI is much more capable of generalizing to multiple languages compared to classical CoT-SFT.

Another seemingly natural configuration would be to receive the question in the target language, perform the thinking in a high-resource pivot language (i.e. English), and then respond in the target language again. This approach has already been investigated by \citep{barua2025long} and \citep{lai2024mcot}, which both concluded that this performs worse than performing CoT in the target language. We therefore did not investigate this further here.

\subsection{Number of thinking Tokens: CoT-SFT vs. CODI}
\label{sec:number_of_cot_tokens}
One of the main benefits of CODI is that it requires much fewer thinking tokens compared to classical CoT-SFT \citep{shen2025codicompressingchainofthoughtcontinuous}. This becomes obvious in table \ref{tab:token_comparison} where, even when trained on all target languages ("Multi-Lingual-with-Urdu" setup), standard latent reasoning achieves compression ratios of roughly $29\times$ and $50\times$.
\begin{table}[h]
\centering
\caption{Efficiency comparison between explicit and continuous reasoning. Average CoT lengths are calculated across all 5 test languages using the \textit{Multi-Lingual-with-Urdu} training setup (see Tables \ref{tab:cot_length_gsm8k_aug_nl} \& \ref{tab:cot_length_commonsenseqa} in the appendix).}
\label{tab:token_comparison}
\begin{tabular}{lccc}
\toprule
\textbf{Dataset} & \textbf{CoT-SFT (Avg Tokens)} & \textbf{CODI (Latent Tokens)} & \textbf{Compression Ratio} \\
\midrule
GSM8K-Aug-NL & $\sim$176 & 6 & $\sim$29$\times$ \\
CommonsenseQA & $\sim$299 & 6 & $\sim$50$\times$ \\
\bottomrule
\end{tabular}
\end{table}

\section{Conclusion}

Performing CoT reasoning in continuous latent space instead of token space offers several advantages, such as requiring fewer thinking tokens \citep{shen2025codicompressingchainofthoughtcontinuous} and performing several thoughts in parallel \citep{coconut}. In this work we found empirical evidence that continuous CoT also learns more language-agnostic representations which allows it to generalize better to new languages and perform better in low-resource languages. 

In future work, we plan to directly analyze the learned continuous representations to verify their language-agnostic properties. Additionally, we aim to scale this investigation to larger models and a broader range of datasets to determine if these findings hold across different architectures and domains.

\section{Acknowledgments}
This research has been funded by the Federal Ministry of Education and Research of Germany and the state of North-Rhine Westphalia as part of the Lamarr Institute for Machine Learning and Artificial Intelligence.

\bibliography{iclr2026_conference}
\bibliographystyle{iclr2026_conference}

\appendix
\section{Appendix}

\begin{table}[h!]
    \centering
    \sisetup{table-format=2.2}
    \caption{Accuracy (\%) on GSM8K-Aug-NL (test split): CODI vs. COT-SFT across training setups.}
    \label{tab:gsm8k_aug_nl_gsm8kaug_codi_vs_cotsft}
    \begin{tabular}{lll SS}
        \toprule
        \multirow{2}{*}{\textbf{Training Dataset}} &
        \multirow{2}{*}{\textbf{Train Setup}} &
        \multirow{2}{*}{\textbf{Test Language}} &
        \multicolumn{2}{c}{\textbf{Accuracy (\%)}} \\
        \cmidrule(lr){4-5}
        & & & \textbf{CODI} & \textbf{COT-SFT} \\
        \midrule

        \multirow{10}{*}{GSM8K-Aug-NL}
            & \multirow{5}{*}{Multi-Lingual}
                & German  & 44.76 & 43.20 \\
            & & French  & 45.19 & 45.07 \\
            & & Chinese & 43.75 & 35.60 \\
            & & English & 46.47 & 51.20 \\
            & & Urdu    & 12.28 & 6.84 \\

        \addlinespace[0.3em]
        \cmidrule(lr){2-5}
        \addlinespace[0.2em]

            & \multirow{5}{*}{Multi-Lingual-with-Urdu}
                & German  & 47.01 & 45.26 \\
            & & French  & 44.58 & 48.98 \\
            & & Chinese & 44.27 & 39.27 \\
            & & English & 47.93 & 52.16 \\
            & & Urdu    & 35.03 & 30.86 \\
        \bottomrule
    \end{tabular}
\end{table}

\begin{table}[h!]
    \centering
    \sisetup{table-format=2.2}
    \caption{Accuracy (\%) on CommonSenseQA (test split) across test languages for different training setups (CODI vs. COT-SFT).}
    \label{tab:commonsenseqa_codi_vs_cotsft}
    \begin{tabular}{lll SS}
        \toprule
        \multirow{2}{*}{\textbf{Training Dataset}} &
        \multirow{2}{*}{\textbf{Train Setup}} &
        \multirow{2}{*}{\textbf{Test Language}} &
        \multicolumn{2}{c}{\textbf{Accuracy (\%)}} \\
        \cmidrule(lr){4-5}
        & & & \textbf{CODI} & \textbf{COT-SFT} \\
        \midrule

        \multirow{10}{*}{CommonSenseQA}
            & \multirow{5}{*}{Multi-Lingual}
                & German  & 58.72 & 49.71 \\
            & & French  & 57.65 & 49.39 \\
            & & Chinese & 57.66 & 51.52 \\
            & & English & 71.42 & 62.49 \\
            & & Urdu    & 35.95 & 23.01 \\

        \addlinespace[0.3em]
        \cmidrule(lr){2-5}
        \addlinespace[0.2em]

            & \multirow{5}{*}{Multi-Lingual-with-Urdu}
                & German  & 56.43 & 50.70 \\
            & & French  & 56.84 & 51.92 \\
            & & Chinese & 56.67 & 50.53 \\
            & & English & 68.96 & 63.61 \\
            & & Urdu    & 41.20 & 34.73 \\
        \bottomrule
    \end{tabular}
\end{table}

\begin{table}[h!]
    \centering
    \sisetup{table-format=2.2}
    \caption{Base model accuracy (\%) on GSM8K-Aug-NL by test language (independent of fine-tuning setup).}
    \label{tab:appendix_base_gsm8k_aug_by_language}
    \begin{tabular}{l S}
        \toprule
        \textbf{Test Language} & {\textbf{Base Accuracy (\%)}} \\
        \midrule
        German  & 26.84 \\
        French  & 25.32 \\
        Chinese & 23.50 \\
        English & 36.01 \\
        Urdu    & 7.66  \\
        \bottomrule
    \end{tabular}
\end{table}

\begin{table}[h!]
    \centering
    \sisetup{table-format=2.2}
    \caption{Base model accuracy (\%) on CommonSenseQA (test split) by test language (independent of fine-tuning setup).}
    \label{tab:appendix_commonsenseqa_base_by_language}
    \begin{tabular}{l S}
        \toprule
        \textbf{Test Language} & {\textbf{Base Accuracy (\%)}} \\
        \midrule
        German  & 33.09 \\
        French  & 39.89 \\
        Chinese & 41.28 \\
        English & 44.49 \\
        Urdu    & 26.54 \\
        \bottomrule
    \end{tabular}
\end{table}

\section{Training Hyperparameters}
\label{app:hyperparameters}

\begin{table}[h]
\centering
\caption{Training hyperparameters for CODI and CoT-SFT. All parameters are kept constant across both GSM8k-Aug-NL and CommonsenseQA datasets to ensure fair comparison.}
\label{tab:hyperparameters}
\small
\begin{tabular}{lc}
\toprule
\textbf{Hyperparameter} & \textbf{Value} \\
\midrule
\multicolumn{2}{l}{\textit{Shared Parameters (CODI \& CoT-SFT)}} \\
\midrule
Learning Rate & $8 \times 10^{-4}$ \\
LR Scheduler & Cosine \\
Warmup Ratio & 0.03 \\
Weight Decay & 0.1 \\
Max Gradient Norm & 2.0 \\
Training Epochs & 10 \\
Max Sequence Length & 512 \\
Precision & BF16 \\
Random Seed & 11 \\
Effective Batch Size & 128 \\
\midrule
\multicolumn{2}{l}{\textit{LoRA Configuration}} \\
\midrule
Rank ($r$) & 128 \\
Alpha ($\alpha$) & 32 \\
Dropout & 0.1 \\
Target Modules & q, k, v, o, gate, up, down \\
\midrule
\multicolumn{2}{l}{\textit{CODI-Specific Parameters}} \\
\midrule
Number of Latents & 6 \\
Projection Dimension & 2048 \\
Distillation Loss Factor & 20 \\
Normalize Distillation Loss & Yes \\
\bottomrule
\end{tabular}
\vspace{0.5em}

\end{table}

\begin{table}[t]
\centering
\caption{Average Chain-of-Thought (CoT) length (tokens) on GSM8K\_AUG\_NL across training setups and test languages (CoT-SFT).}
\label{tab:cot_length_gsm8k_aug_nl}
\small
\begin{tabular}{llcc}
\toprule
\textbf{Training Dataset} & \textbf{Train Setup} & \textbf{Test Language} & \textbf{Avg CoT Length (tokens)} \\
\midrule
\multirow{15}{*}{GSM8K\_AUG\_NL}
& \multirow{5}{*}{English}
& German  & 172.2 \\
&  & French  & 161.6 \\
&  & Chinese & 171.4 \\
&  & English & 155.4 \\
&  & Urdu    & 512.0 \\
\cmidrule(lr){2-4}
& \multirow{5}{*}{Multi-Lingual}
& German  & 239.6 \\
&  & French  & 211.7 \\
&  & Chinese & 175.9 \\
&  & English & 156.3 \\
&  & Urdu    & 503.0 \\
\cmidrule(lr){2-4}
& \multirow{5}{*}{Multi-Lingual-with-Urdu}
& German  & 187.9 \\
&  & French  & 186.8 \\
&  & Chinese & 151.0 \\
&  & English & 154.0 \\
&  & Urdu    & 199.3 \\
\bottomrule
\end{tabular}
\end{table}

\begin{table}[t]
\centering
\caption{Average Chain-of-Thought (CoT) length (tokens) on CommonsenseQA across training setups and test languages (CoT-SFT).}
\label{tab:cot_length_commonsenseqa}
\small
\begin{tabular}{llcc}
\toprule
\textbf{Training Dataset} & \textbf{Train Setup} & \textbf{Test Language} & \textbf{Avg CoT Length (tokens)} \\
\midrule
\multirow{15}{*}{CommonsenseQA}
& \multirow{5}{*}{English}
& German  & 455.1 \\
&  & French  & 329.1 \\
&  & Chinese & 203.8 \\
&  & English & 205.5 \\
&  & Urdu    & 512.0 \\
\cmidrule(lr){2-4}
& \multirow{5}{*}{Multi-Lingual}
& German  & 296.8 \\
&  & French  & 286.7 \\
&  & Chinese & 228.4 \\
&  & English & 208.0 \\
&  & Urdu    & 512.0 \\
\cmidrule(lr){2-4}
& \multirow{5}{*}{Multi-Lingual-with-Urdu}
& German  & 303.5 \\
&  & French  & 294.7 \\
&  & Chinese & 215.2 \\
&  & English & 207.6 \\
&  & Urdu    & 476.0 \\
\bottomrule
\end{tabular}
\end{table}

\section{Dataset Construction Details}
\label{app:dataset_construction}

\subsection{GSM8k-Aug-NL Multilingual Dataset}

We construct multilingual versions of GSM8k-Aug-NL by translating the English dataset into German, French, Chinese, and Urdu. Each translated sample maintains a reference to its English original, enabling cross-language deduplication to ensure zero overlap, each mathematical problem appears in exactly one language.

For the five-language configuration, we prioritize Urdu (4,619 samples) to maximize low-resource coverage, then distribute remaining problems across high-resource languages based on translation quality.

\begin{table}[h]
\centering
\caption{GSM8k-Aug-NL multilingual dataset statistics.}
\label{tab:gsm8k_stats}
\small
\begin{tabular}{lrr}
\toprule
\textbf{Language} & \textbf{Samples} & \textbf{\% of Total} \\
\midrule
English & 95,131 & 24.5\% \\
German & 96,139 & 24.8\% \\
French & 95,868 & 24.7\% \\
Chinese & 96,178 & 24.8\% \\
Urdu & 4,619 & 1.2\% \\
\midrule
\textbf{Total} & \textbf{387,935} & 100\% \\
\bottomrule
\end{tabular}
\end{table}

\subsection{CommonsenseQA Multilingual Dataset}

We apply the same methodology to CommonsenseQA, ensuring zero overlap across languages.

\begin{table}[h]
\centering
\caption{CommonsenseQA multilingual dataset statistics.}
\label{tab:csqa_stats}
\small
\begin{tabular}{lrr}
\toprule
\textbf{Language} & \textbf{Samples} & \textbf{\% of Total} \\
\midrule
English & 1,940 & 20.0\% \\
German & 1,940 & 20.0\% \\
French & 1,940 & 20.0\% \\
Chinese & 1,940 & 20.0\% \\
Urdu & 1,940 & 20.0\% \\
\midrule
\textbf{Total} & \textbf{9,700} & 100\% \\
\bottomrule
\end{tabular}
\end{table}

\end{document}